\documentclass{article}
\usepackage{spconf,amsmath,graphicx,hyperref}

\makeatletter
\renewcommand\section{%
  \@startsection{section}{1}{\z@}%
    {1.0ex \@plus .3ex \@minus .2ex}%   % before-skip (was larger in article.cls)
    {0.6ex \@plus .2ex}%                % after-skip  (space to following text)
    {}}

\renewcommand\subsection{%
  \@startsection{subsection}{2}{\z@}%
    {0.8ex \@plus .3ex \@minus .2ex}%
    {0.5ex \@plus .2ex}%
    {}}

\renewcommand\subsubsection{%
  \@startsection{subsubsection}{3}{\z@}%
    {0.6ex \@plus .2ex \@minus .1ex}%
    {0.4ex \@plus .2ex}%
    {}}
\makeatother

\usepackage{todonotes}
\usepackage{amsfonts}
\usepackage{booktabs}
\usepackage{adjustbox}

\newcommand\glass{\texttt{GLASS}}

\title{Encoding Emotion through Self-Supervised \\Eye Movement Reconstruction}

\name{Marcus Ma$^1$, Jordan Prescott$^1$, Emily Zhou$^1$, Tiantian Feng$^1$,\\\em{Kleanthis Avramidis$^1$, Gabor Mihaly Toth$^2$, Shrikanth Narayanan$^1$}}
\address{$^1$University of Southern California, $^2$University of Luxembourg}

\usepackage{fancyhdr}
\fancypagestyle{firstpage}{
\fancyhf{}
\fancyfoot[C]{\footnotesize Copyright 2026 IEEE. Published in ICASSP 2026 - 2026 IEEE International Conference on Acoustics, Speech and Signal Processing (ICASSP), scheduled for 3-8 May 2026 in Barcelona, Spain. Personal use of this material is permitted. However, permission to reprint/republish this material for advertising or promotional purposes or for creating new collective works for resale or redistribution to servers or lists, or to reuse any copyrighted component of this work in other works, must be obtained from the IEEE. Contact: Manager, Copyrights and Permissions / IEEE Service Center / 445 Hoes Lane / P.O. Box 1331 / Piscataway, NJ 08855-1331, USA. Telephone: + Intl. 908-562-3966.}

}
\begin{document}
\ninept
\maketitle

\begin{abstract}
The relationship between emotional expression and eye movement is well-documented, with literature establishing gaze patterns are reliable indicators of emotion. However, most studies utilize specialized, high-resolution eye-tracking equipment, limiting the potential reach of findings. We investigate how eye movement can be used to predict multimodal markers of emotional expression from naturalistic, low-resolution videos. We utilize a collection of video interviews from the USC Shoah Foundation's Visual History Archive with Holocaust survivors as they recount their experiences in the Auschwitz concentration camp. Inspired by pretraining methods on language models, we develop a novel gaze detection model that uses self-supervised eye movement reconstruction that can effectively leverage unlabeled video. We use this model's encoder embeddings to fine-tune models on two downstream tasks related to emotional expression. The first is aligning eye movement with directional emotion estimates from speech. The second task is using eye gaze as a predictor of three momentary manifestations of emotional behaviors: laughing, crying/sobbing, and sighing. We find our new model is predictive of emotion outcomes and observe a positive correlation between pretraining performance and emotion processing performance for both experiments. We conclude self-supervised eye movement reconstruction is an effective method for encoding the affective signal they carry.
\end{abstract}

\section{Introduction}
\label{sec:intro}
The eyes have long been called windows to the soul; but beyond metaphor, extensive research studies have shown that eye movement is closely associated with emotional state \cite{lim_2020, odwyer2017}. Emotion processing manifests subconsciously in subtle shifts to gaze patterns and eye fixations, and recent approaches in eye processing reveal that it is even capable of identifying markers of suicidality \cite{avramidis_2025}. High-quality eye tracking data has traditionally been collected via specialized equipment designed specifically for pupil tracking and corneal reflection or via electro-oculography sensors; however, this inherently limits the size and scope of data collection, and its broader applicability.

While specialized eye tracking data will always be useful, it is expensive, time-consuming, and infeasible to collect at a large scale. In this work, we investigate if eye movement information derived from naturalistic settings -- in our case, sit-down, low-resolution videos of interviews -- are able to convey meaningful rich affective signal. We process a large-scale video dataset of Holocaust survivor interviews containing thousands of hours of eye behavior data. This dataset captures intense ranges of emotional display, with thousands of instances of sobbing, laughing, and crying in their videos. Despite the available low-quality data resolution (30 FPS at 320p), we show that this affective signal can still be predicted by eye movement alone. To utilize this large dataset, and taking inspiration from approaches in NLP, we design an eye gaze model tasked with self-supervised movement reconstruction. We pretrain our model, \glass{}, to autoregressively predict future eye movement in a sequence-to-sequence encoder-decoder architecture with Transformer blocks. After pretraining, we take the encoder portion of \glass{} and add a downstream emotion model head finetuned on our limited emotion data in two experiments. We find \glass{} outperforms baselines on both experiments and discover several other findings, such as that with the same architecture and data, \glass{} model performance improves by just modifying the pretraining task of gaze reconstruction to target a larger forecast. Our findings illustrate that self-supervised eye movement reconstruction is an effective approach for encoding affect from gaze data\footnote{Code available at \href{https://github.com/mamarcus64/GLASS}{github.com/mamarcus64/GLASS}.}.

\begin{figure}[t]
    \centering
    \includegraphics[width=\linewidth]{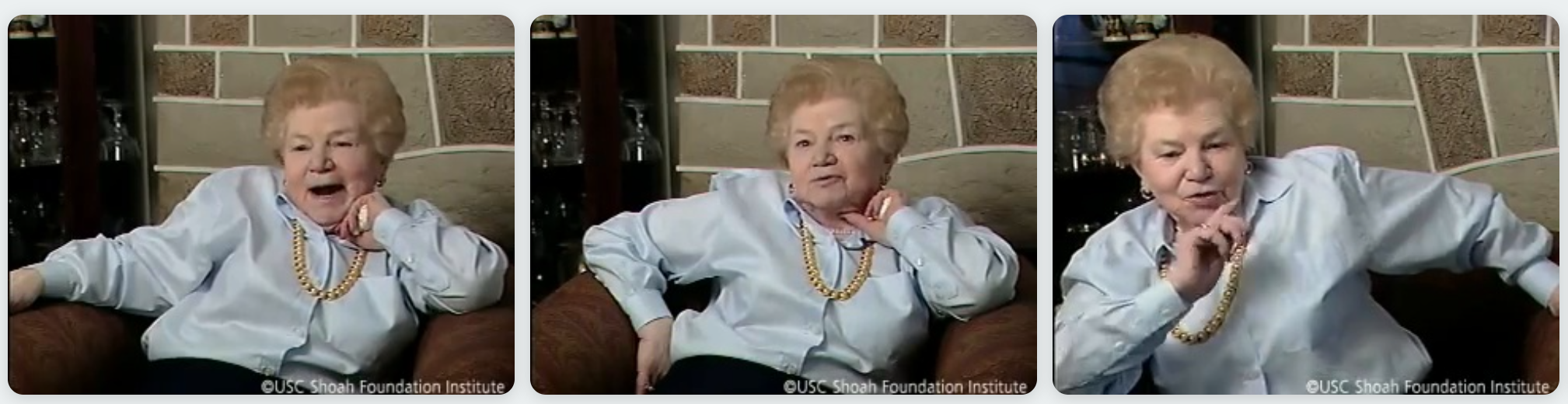}
    \vspace{-0.5cm}
    \caption{Example frames of a sample interview with a Holocaust survivor from the VOICES dataset.}
    \vspace{-0.2cm}
    \label{fig:interview_example}
\end{figure}

\section{Related Work}
\label{sec:related_work}
\subsection{Eye Gaze and Affect}
Eye movement has been shown to be a salient predictor of experienced affect \cite{lanata2011, soleymani2012, odwyer2017, edinger2024}. In particular, eye gaze was found to be a stronger predictor of affect than physiological signals in contexts where emotions were induced with visual stimuli \cite{soleymani2012}. Features derived from eye movement can also effectively identify different levels of arousal and valence \cite{lanata2011} and provide enhanced discrimination when combined with speech features in continuous affect prediction \cite{odwyer2017}. However, the majority of such works have focused on data collected in lab settings, where lighting conditions, data collection quality, and emotional stimuli can are carefully controlled.

\begin{figure*}[t]
    \centering
    \includegraphics[width=\textwidth]{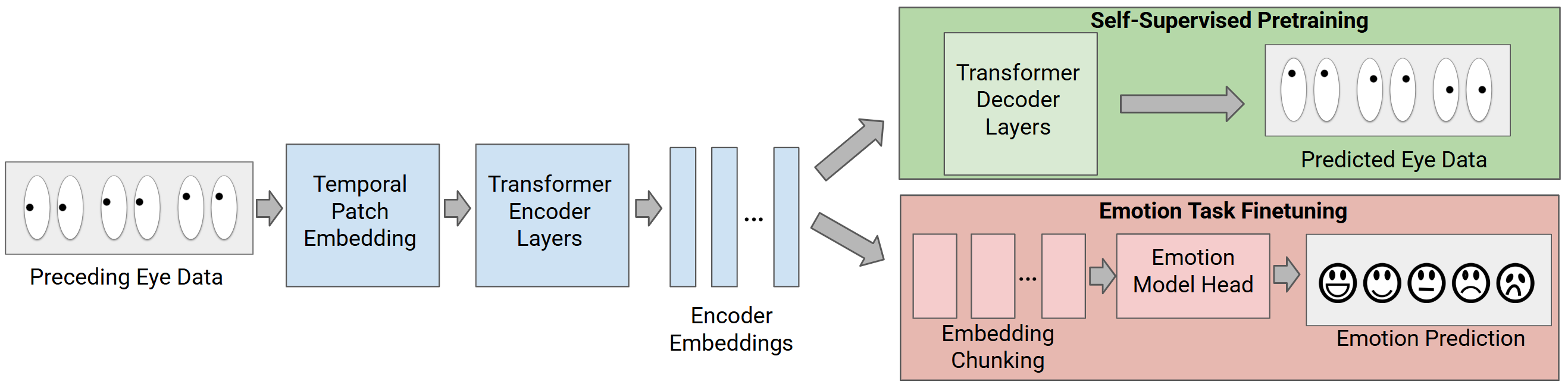}
    \caption{Architecture of \glass{}. Pretraining starts with an encoder-decoder eye gaze prediction task and for downstream prediction the decoder is replaced with an emotion model head.}
    \label{fig:glass}
\end{figure*}

\subsection{Self-Supervised Learning}
Self-supervised learning (SSL) enables the learning of generalizable representations of input data without requiring labeled data, which can be a constraint in affective computing research. SSL has been successfully applied to affective computing across many modalities, including physiological signals \cite{sarkar2022, wu2024}, audio \cite{nimitsurachat2023}, and facial video \cite{sun2025}, achieving better performance than traditional supervised approaches. Eye gaze in more naturalistic settings is still an underexplored modality in affective computing. Motivated by past works demonstrating the effectiveness of eye gaze in affect detection, we explore SSL techniques applied to the more challenging modality of noisy, camera-based naturalistic eye gaze. 

\section{Dataset}
\label{sec:dataset}
We use selected subset of videos from the USC Shoah Foundation's Visual History Archive\footnote{See \href{https://vha.usc.edu/home}{hva.usc.edu/home}. Though we process 978 subjects, there are over 50,000 testimonies available from the Holocaust and other global atrocities.}, a collection of sit-down, semi-structured interviews where participants recount lived experiences of several global atrocities. We narrow our scope to 3997 videos from 978 Auschwitz concentration camp survivors (526 men; 452 women) held captive from 1940 to 1945 with interviews recorded between 1994 and 1998. Each video is 30 FPS and 320x240 pixel resolution and approximately 30 minutes long, with about 4 videos per interviewee. These interviews were professionally annotated with time-aligned transcripts, providing us with a unique opportunity to analyze narratives about these traumatic experiences across three rich modalities: audio, video, and text.

\subsection{Eye Gaze Extraction}
We use OpenFace 2.0 \cite{baltrusaitis_2018} to extract eye gaze measures from the video recordings. OpenFace outputs frame-level data of the XYZ gaze direction for each eye, blinks, and facial expression landmarks.

\begin{table}
    \centering
    \begin{tabular}{lccc}
    \toprule
        & \textbf{Pretraining} & \textbf{Exp. 1} & \textbf{Exp. 2} \\
        \midrule
        Labels & Gaze & VAD & Behaviors \\
        \# Videos & 3997 & 3979 & 1926\\
        \# Subjects & 978 & 986 & 715\\
        \# Examples & 3.4M & 54k & 7.4k\\
        \bottomrule
    \end{tabular}
    \caption{Data for \glass{} pretraining, Exp. 1, and Exp. 2. VAD indicates valence, arousal, and dominance.}
    \vspace{-3.5mm}
    \label{tab:datasets}
\end{table}

\subsection{Experiment Data}

We conduct two experiments to investigate how eye gaze can predict affect: a regression task directly predicting emotion values under the Valence-Arousal-Dominance model \cite{mehrabian_1974} (Exp. 1) and a classification task predicting specific behaviors associated with highly salient emotions (Exp. 2). For both experiments, we bootstrap 5 seeds of an 80-20 train-test split.

\begin{figure}
    \centering
    \includegraphics[width=0.96\linewidth]{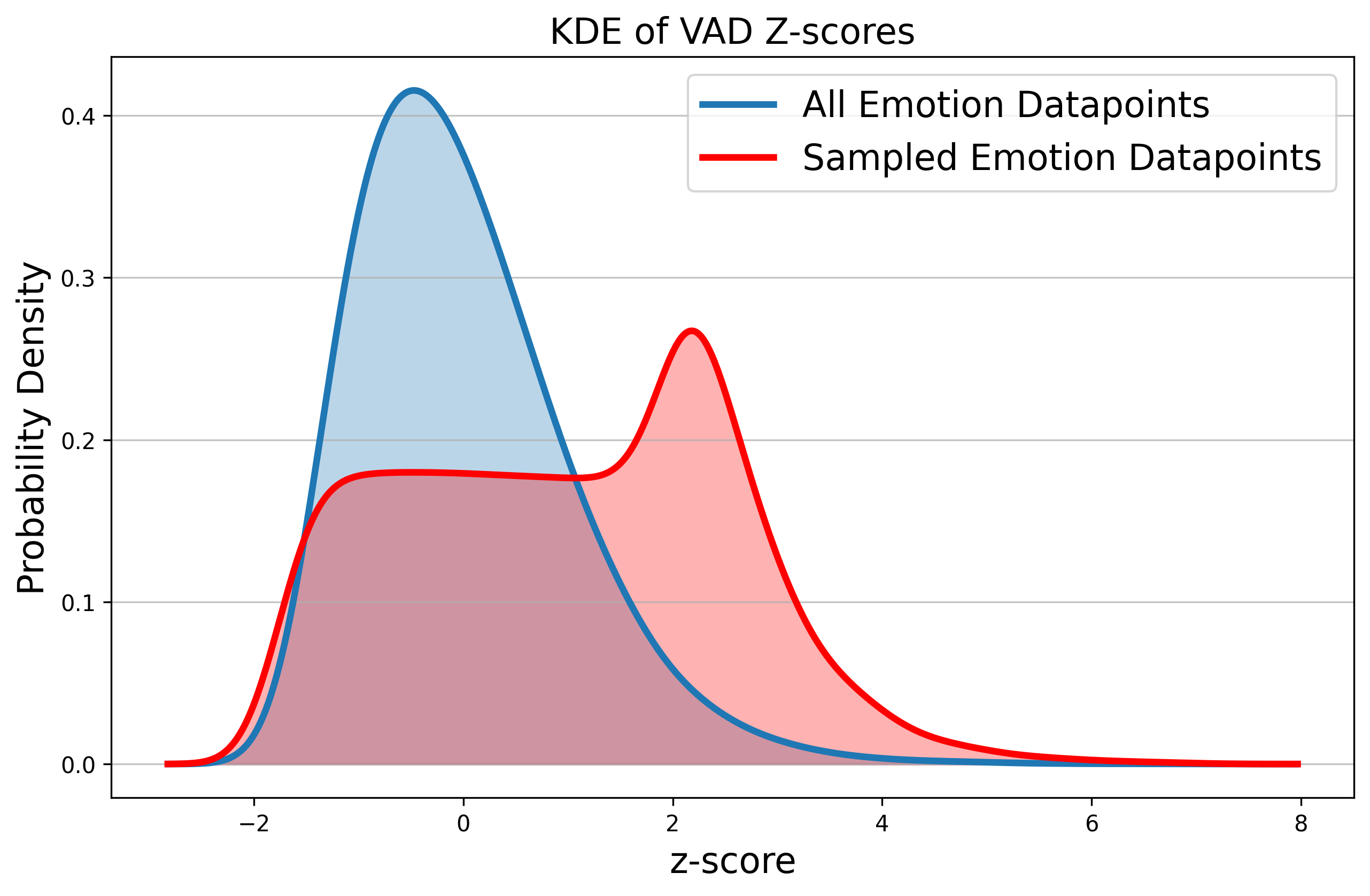}
    \vspace{-0.2cm}
    \caption{Kernel density estimates (KDEs) of emotional state distributions as a function of the z-score of their Euclidean distance from the dataset mean, comparing our original audio model with our modified sampling. VAD
indicates valence, arousal, and dominance.}
    \label{fig:vad_plot}
\end{figure}

\begin{figure*}[t]
    \centering
    \includegraphics[width=\textwidth]{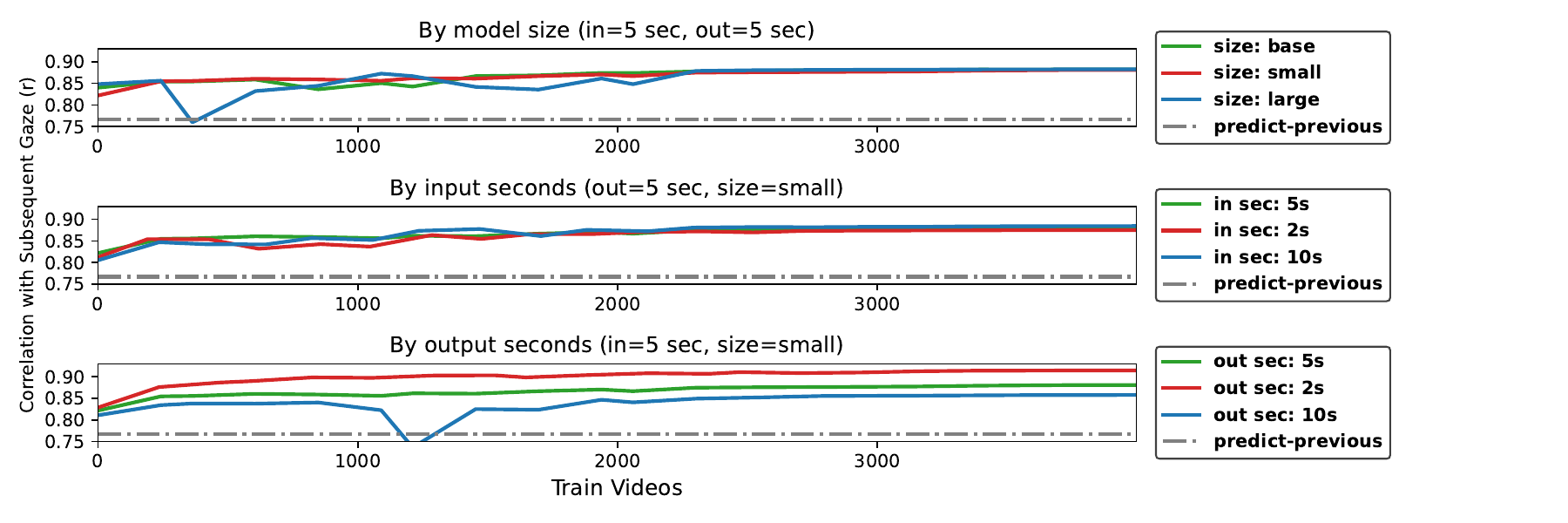}
    \vspace{-0.3in}
    \caption{Performance of \glass{} on the self-supervised task of subsequent gaze prediction on a held-out validation set. Three different conditions for \glass{} are tested: model size (top), number of input seconds received (middle), and number of subsequent seconds predicted (bottom). The predict-previous baseline predicts the next frame to be the current frame after teacher-forcing. All \glass{} models predict autoregressively without teacher-forcing.}
    \label{fig:glass_train}
\end{figure*}

\subsubsection{Experiment 1 - VAD Regression} 

To label the videos with emotion data, we use the challenge-winning emotion prediction pipeline from the Speech Emotion Recognition in Naturalistic Conditions Challenge at INTERSPEECH-2025, described in \cite{lertpetchpun25_interspeech} and are based on the Vox-Profile models~\cite{feng2025vox}. These pipelines use speaker audio to estimate emotion output from two ASR systems, Whisper and WavLM. Specifically, the emotional attribute prediction models output arousal, valence, and dominance on a scale between 0 and 1. The models are fine-tuned with Whisper-Large~\cite{radford2023robust} and WavLM-Large~\cite{chen2022wavlm} backbones on the MSP-Podcast dataset~\cite{lotfian2017building}, and we ensemble the outputs from these two models to obtain the final emotional attribute predictions. For this dataset, we use the transcripts to collect windows when the interviewee is speaking and calculate one VAD value per sentence spoken, approximately 5-10 seconds in length. Both Whisper-based~\footnote{\href{https://huggingface.co/tiantiaf/whisper-large-v3-msp-podcast-emotion-dim}{huggingface.co/tiantiaf/whisper-large-v3-msp-podcast-emotion-dim}} and WavLM-based~\footnote{\href{https://huggingface.co/tiantiaf/wavlm-large-msp-podcast-emotion-dim}{huggingface.co/tiantiaf/wavlm-large-msp-podcast-emotion-dim}} models are available on HuggingFace.

Upon examination of the VAD output distribution, the majority of VAD outputs were centered closely around the mean of $[0.328, 0.410, 0.388]$, significantly below the ``neutral'' emotion mean of $[0.5, 0.5, 0.5]$. To increase dataset diversity, we decide to upsample VAD labels that lie at least 2 standard deviations away from the VAD mean to a ratio of 1/3 (shown in Fig.~\ref{fig:vad_plot}). For the task itself, models have access to the eye movement data of the five seconds \textbf{leading up to} the label, which we sample with a 3-second stride over each corresponding VAD sentence window for a total of 54,374 windows. We evaluate model performance on two metrics: the average of sample-wise Mean Absolute Error (MAE), and the test-set-wise Pearson correlation (r) between predicted and ASR-generated VAD.

\subsubsection{Experiment 2 - Behavior Classification}
The transcripts provided by the Shoah Foundation include not only text output but also several non-verbal behavior markers. In particular, we examine the markers for three emotionally-relevant behaviors: crying/sobbing, laughing, and sighing. The data had 4,866 instances of laughter, 1,952 sighs, and 1,478 sobs/crying. We report macro-F1 as our performance metric.

\section{Methods}
\label{sec:methods}

\begin{table*}[t]
\centering
\begin{adjustbox}{width=\textwidth}
\begin{tabular}{lccccccccc}
\toprule
 & \multicolumn{3}{c}{Preceding-2-second input} & \multicolumn{3}{c}{Preceding-5-second input} & \multicolumn{3}{c}{Preceding-10-second input} \\
\cmidrule(lr){2-4} \cmidrule(lr){5-7} \cmidrule(lr){8-10}
 & MAE & Pearson's r & F1 & MAE & Pearson's r & F1 & MAE & Pearson's r & F1 \\
\midrule
Statistical Features (eyes only) & $0.110_{\pm 0.01}$ & $0.157_{\pm 0.05}$ & $0.271_{\pm 0.01}$ & $0.110_{\pm 0.01}$ & $0.226_{\pm 0.03}$ & $0.284_{\pm 0.02}$ & $0.110_{\pm 0.01}$ & $0.223_{\pm 0.03}$ & $0.290_{\pm 0.02}$ \\

Statistical Features (eyes + face) & $0.110_{\pm 0.01}$ & $0.167_{\pm 0.05}$ & $0.272_{\pm 0.02}$ & $0.110_{\pm 0.01}$ & $0.228_{\pm 0.03}$ & $0.294_{\pm 0.02}$ & $0.110_{\pm 0.01}$ & $0.225_{\pm 0.03}$ & $0.303_{\pm 0.03}$ \\

Eye-Gaze Temporal CNN & ${\textbf{0.105}}_{\pm 0.01}$ & $0.166_{\pm 0.05}$ & $0.267_{\pm 0.02}$ & ${\textbf{0.106}}_{\pm 0.01}$ & $0.157_{\pm 0.06}$ & $0.285_{\pm 0.02}$ & ${\textbf{0.108}}_{\pm 0.01}$ & $0.148_{\pm 0.05}$ & $0.293_{\pm 0.02}$ \\

\midrule
\glass{} predicting 2 seconds & $0.152_{\pm 0.03}$ & $0.230_{\pm 0.04}$ & $0.332_{\pm 0.02}$ & $0.161_{\pm 0.04}$ & $0.228_{\pm 0.05}$ & $0.341_{\pm 0.02}$ & $0.154_{\pm 0.02}$ & $0.167_{\pm 0.05}$ & $0.326_{\pm 0.01}$ \\

\glass{} predicting 5 seconds & $0.123_{\pm 0.01}$ & $0.283_{\pm 0.04}$ & ${\textbf{0.367}}_{\pm 0.02}$ & $0.122_{\pm 0.00}$ & $0.284_{\pm 0.03}$ & ${\textbf{0.361}}_{\pm 0.02}$ & $0.125_{\pm 0.00}$ & $0.285_{\pm 0.04}$ & ${\textbf{0.352}}_{\pm 0.02}$ \\

\glass{} predicting 10 seconds & $0.119_{\pm 0.01}$ & ${\textbf{0.285}}_{\pm 0.04}$ & $0.356_{\pm 0.01}$ & $0.122_{\pm 0.00}$ & ${\textbf{0.294}}_{\pm 0.03}$ & $0.348_{\pm 0.01}$ & $0.122_{\pm 0.00}$ & ${\textbf{0.297}}_{\pm 0.03}$ & $0.347_{\pm 0.02}$ \\

\bottomrule
\end{tabular}
\end{adjustbox}

\caption{Results from the Exp. 1 (MAE and Pearson's r) and the Exp. 2 (F1) for baselines and \glass{} small. Each version of \glass{} has the same architecture but was trained to predict a different time frame during the encoder-decoder pretraining. Mean$_{\pm\text{std}}$ over 5 runs.}
\label{tab:results}
\end{table*}

\subsection{Baselines}

We evaluate three baseline models: two statistics-based methods and a CNN approach. For the stats-based methods, for each input dimension (X, Y, and Z for each eye), we calculate mean, std, velocity, acceleration, and cross-input correlation as features. This comprises one stats method (eyes only), and we additionally add these features for blinks, eyebrows, and other facial features for the second one (with both eyes + face). Our CNN baseline comprises a 1D Temporal CNN over the 6-channel gaze time series with ReLU, BatchNorm, and Dropout, then global pooling and a MLP head.

\subsection{\glass}

We propose a novel model, \textbf{G}aze-based \textbf{L}earning of \textbf{A}ffect through \textbf{S}elf-\textbf{S}upervision (\glass{}).
In our dataset, the amount of unlabeled eye gaze data greatly exceeds the amount of labeled emotion data. Taking inspiration from pretraining approaches in NLP, we develop \glass{} in two phases: first, a base model pretrained on the self-supervised task of predicting future eye movement using an encoder-decoder. Second, we train an emotion head mapping the learned encoder embeddings to predict emotion.

\subsubsection{Training the Encoder-Decoder}
We model gaze prediction as a temporal sequence-to-sequence prediction task. Given input eye data spanning $D$ dimensions over $T_i$ frames attempting to predict the next $T_o$ frames, \glass{} learns the mapping $
    \mathbb{R}^{T_i \times D} \rightarrow \mathbb{R}^{T_o \times D}$.
For our OpenFace data containing the XYZ coordinates of two eyeballs, $D=6$, and for predicting 5 seconds of eye movement with 30 FPS resolution, we have $T_i=T_o=150$. Following \cite{dosovitskiy_2021}, we combine consecutive frames into non-overlapping patches then embed each patch via a linear embedding $\mathbb{R}^{P\times D}\rightarrow \mathbb{R}^{P\cdot d}$, where $d$ is our embedding dimension and $P$ is the patch size. We supply positional information to these embeddings via RoPE \cite{su_2023}, which rotates the embeddings based on their relative sequence position. Our encoder is a stack of $L_e$ standard self-mapping Transformer blocks operating on $\mathbb{R}^{T_i/P\times d}$. Our decoder is a series of $L_d$ Transformer blocks with cross-attention over the encoder hidden states that auto-regressively recovers patch-level embeddings $f_t \in \mathbb{R}^d$. Each recovered patch is then mapped back to gaze features ($\mathbb{R}^{P \times D}$) via a linear mapping; the final predicted gaze sequence is achieved by concatenating these features. We train three model sizes: small, base, and large, each with different hidden dimensions and number of Transformer blocks.

We train \glass{} by taking all frames with eye data from all 3,997 videos across 978 subjects, reserving 49 subjects for validation, and extracting consecutive windows of 5 seconds (150 frames) of input and the subsequent 5 seconds as output. We use a stride of 151 frames to ensure no individual patch appears twice among windows. We utilize a scheduled sampling regimen \cite{bengio_2015} for decoding such that at the beginning of training, previously decoded values will be teacher-forced (i.e., replaced with the ground-truth value) with 100\% probability, which linearly reduces to 0\% probability (i.e., entirely autoregressive) when 60\% of training is complete. Validation is always performed entirely autoregressively. We use a joint loss of both coordinate and velocity accuracy via $\mathcal{L} = \mathcal{L}_c + \lambda \cdot \mathcal{L}_v$ with $\lambda=0.2$. Both terms utilize Huber loss \cite{huber_1964} over each frame, with coordinate and velocity accuracy corresponding to $|\hat{y_t} - y_t|$ and $|\Delta\hat{y_t} - \Delta y_t|$, respectively. We utilize AdamW \cite{loshchilov_2019} with a cosine-decayed learning rate of 3e-4 over 3000 warmup steps and a weight decay of 1e-4. Pretraining results are presented in Fig.~\ref{fig:glass_train}. We include a "predict-previous" baseline where future predictions are set to be the current frame after teacher-forcing, which by itself achieves an impressive correlation of 0.767 on a held-out validation set. However, all \glass{} models are able to significantly beat out this baseline even though unlike this baseline, all output is determined autoregressively without teacher forcing.

\subsubsection{Finetuning the Emotion Head}
After pretraining, we finetune \glass{} by replacing the decoder with an emotion head. We calculate features from the encoder embeddings (embedding features and their first and second order derivative estimates) and combine them into non-overlapping chunks to pass into our emotion head. We evaluate four different temporal models as the emotion head: a time-averaged MLP; a Temporal Convolutional Network; a Gated Recurrent Unit; and a Transformer. We also test chunk sizes for 0.5, 1, 2, and 4 seconds. In total, all models were trained on a single A40 GPU on a secure Linux cluster for approximately 200 petaFLOPs and 6 hours.

\section{Results}
\label{sec:results}
Results for Exp. 1 and Exp. 2 are presented in Table~\ref{tab:results}. We highlight two general findings and three \glass{}-specific insights:
\subsection{Low-quality eye data still holds rich emotion information.}
The best values across all models for MAE (0.105) and Pearson's r (0.297) are all significantly higher than chance, which is not trivial given that our input and output data come from different modalities of video and audio with a tenfold reduction of quality in both resolution and frame rate compared to industry standards\footnote{For example, the \href{ https://www.sr-research.com/wp-content/uploads/2017/11/eyelink-1000-plus-specifications.pdf}{EyeLink 1000} samples at 2000 Hz and can detect eye angle changes within 0.01$^\circ$. Our data is 30 Hz and evaluation of OpenFace 2.0 on the MPIIGaze dataset resulted in a mean gaze error of 9.1$^\circ$ \cite{araluce_2021}.}.

\begin{table}[t]
\centering
\begin{adjustbox}{width=\columnwidth}
\centering
\begin{tabular}{lcccc}
\toprule & MLP & TCN & GRU & Transformer \\
\midrule
0.5 sec chunk & $0.231_{\pm 0.04}$ & $0.198_{\pm 0.04}$ & \textbf{$0.262_{\pm 0.03}$} & $0.258_{\pm 0.03}$ \\
1 sec chunk & $0.240_{\pm 0.04}$ & $0.284_{\pm 0.03}$ & $0.264_{\pm 0.03}$ & $0.253_{\pm 0.03}$ \\
2 sec chunk & $0.242_{\pm 0.04}$ & \textbf{$\textbf{0.285}_{\pm 0.04}$} & $0.258_{\pm 0.03}$ & $0.244_{\pm 0.03}$ \\
4 sec chunk & $0.245_{\pm 0.03}$ & \textbf{$0.275_{\pm 0.04}$} & $0.264_{\pm 0.03}$ & $0.246_{\pm 0.03}$ \\
\bottomrule
\end{tabular}
\end{adjustbox}
\caption{Pearson's r (Exp. 1) for different configurations of $\glass$'s emotion head for the small model. Mean$_{\pm\text{std}}$ over 5 runs.}
\label{tab:chunk_head}
\end{table}

\subsection{Longer gaze sequences are more predictive.}
\label{subsec:longer_gaze}
We display results in Table~\ref{tab:results} for three input lengths: 2, 5, and 10 seconds. Across all models, 5-second and 10-second inputs outperform 2-second input. Interestingly, prior work investigating emotion prediction from eye movement finds windows in the 1 to 2 second range are the most predictive \cite{wang_2018, wu_2021}. However, these approaches use detailed eye-tracking data and capture involuntary micro-movements such as fixations and saccades and controlled stimuli; we hypothesize that when this information is not available due to low resolution or in more natural settings, longer gaze sequences provide a fuller picture on emotional state.

\begin{figure}
    \centering
    \includegraphics[width=\columnwidth]{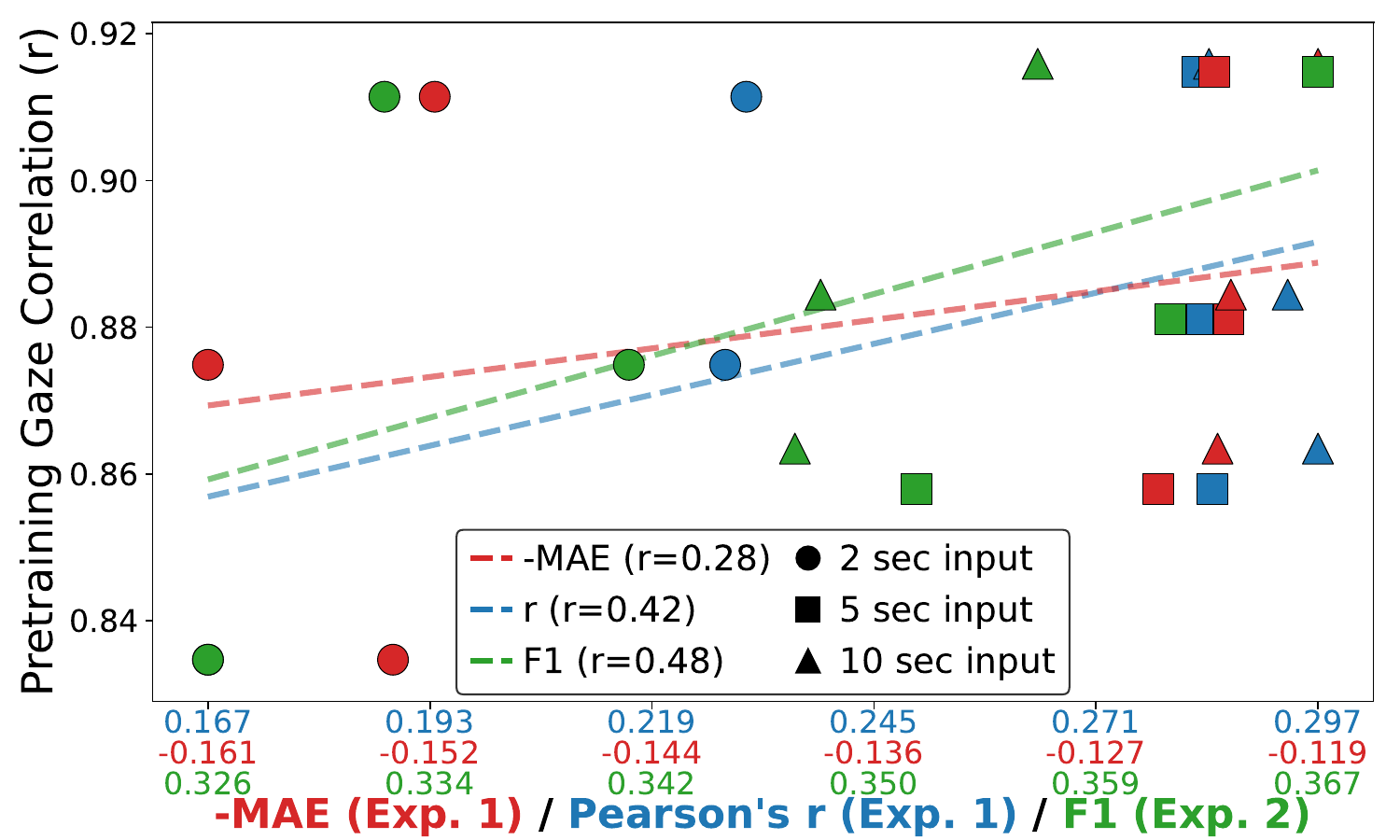}
    \caption{Validation gaze correlation after \glass{}'s self-supervised pretraining vs. Exp. 1 and Exp. 2 metrics. We present negative MAE as lower MAE is more desirable. Point shapes correspond to input time windows.}
    \label{fig:corr_corr}
\end{figure}

\subsection{[\glass{}] Short temporal chunks work best.}
Regarding the 1 to 2 second range input being most effective for deep learning \cite{wang_2018, wu_2021}, we find in Table~\ref{tab:chunk_head} that across all four different emotion head labels, the 1 or 2 second chunk range performs best. Additionally, we find that the worst emotion model head is MLP, which we hypothesize is because it is the only model that does not explicitly consider temporal dynamics.

\subsection{[\glass{}] Longer forecasting improves performance.}
While \glass{} certainly benefits from longer gaze input (\S\ref{subsec:longer_gaze}), we find that performance also increases when it is pretrained to predict a longer time frame. The results in Table~\ref{tab:results} are particularly interesting because the three \glass{} models contain the same architecture and use the same data with the only difference being how many seconds they were trained to predict during pretraining.
\subsection{[\glass{}] Pretraining correlation predicts performance.}
In Fig.~\ref{fig:corr_corr}, we plot the correlation between pretraining performance (validation gaze correlation) and our three experimental metrics and find positive correlations with all three. This confirms that the pretraining task of predicting future eye gaze successfully encodes affective information.

\section{Conclusion}
\label{sec:conclusion}
We show that the rich predictive signal of eye gaze on affect can be extracted even from relatively low-resolution eye data, bolstering the existing body literature on the eye movement--affect interaction to more naturalistic behavior settings. We also introduce \glass{}, an architecture that pretrains on self-supervised eye prediction before finetuning for emotion tasks. We hope to shed more light on the capabilities of low-quality eye data at scale in future work.

\vfill\pagebreak

\section{Acknowledgements}
\noindent This work was supported by the USC Shoah Foundation, the USC Viterbi Graduate Fellowship, the NSF GRFP, the National Science Foundation, the Luxembourg National Research Fund, and the German Research Foundation (Grant C24/ID/18896236/VOICES). 

\bibliographystyle{IEEEbib}
\bibliography{refs}

@INPROCEEDINGS{baltrusaitis_2018,
  author={Baltrusaitis, Tadas and Zadeh, Amir and Lim, Yao Chong and Morency, Louis-Philippe},
  booktitle={2018 13th IEEE International Conference on Automatic Face \& Gesture Recognition (FG 2018)}, 
  title={OpenFace 2.0: Facial Behavior Analysis Toolkit}, 
  year={2018},
  volume={},
  number={},
  pages={59-66},
  doi={10.1109/FG.2018.00019}}

@article{chen2022wavlm,
  title={Wavlm: Large-scale self-supervised pre-training for full stack speech processing},
  author={Chen, Sanyuan and Wang, Chengyi and Chen, Zhengyang and Wu, Yu and Liu, Shujie and Chen, Zhuo and Li, Jinyu and Kanda, Naoyuki and Yoshioka, Takuya and Xiao, Xiong and others},
  journal={IEEE Journal of Selected Topics in Signal Processing},
  volume={16},
  number={6},
  pages={1505--1518},
  year={2022},
  publisher={IEEE}
}

@article{lotfian2017building,
  title={Building naturalistic emotionally balanced speech corpus by retrieving emotional speech from existing podcast recordings},
  author={Lotfian, Reza and Busso, Carlos},
  journal={IEEE Transactions on Affective Computing},
  volume={10},
  number={4},
  pages={471--483},
  year={2017},
  publisher={IEEE}
}

@inproceedings{radford2023robust,
  title={Robust speech recognition via large-scale weak supervision},
  author={Radford, Alec and Kim, Jong Wook and Xu, Tao and Brockman, Greg and McLeavey, Christine and Sutskever, Ilya},
  booktitle={International conference on machine learning},
  pages={28492--28518},
  year={2023},
  organization={PMLR}
}

@article{feng2025vox,
  title={Vox-Profile: A Speech Foundation Model Benchmark for Characterizing Diverse Speaker and Speech Traits},
  author={Feng, Tiantian and Lee, Jihwan and Xu, Anfeng and Lee, Yoonjeong and Lertpetchpun, Thanathai and Shi, Xuan and Wang, Helin and Thebaud, Thomas and Moro-Velazquez, Laureano and Byrd, Dani and others},
  journal={arXiv preprint arXiv:2505.14648},
  year={2025}
}

@inproceedings{lertpetchpun25_interspeech,
  title     = {{Developing a High-performance Framework for Speech Emotion Recognition in Naturalistic Conditions Challenge for Emotional Attribute Prediction}},
  author    = {{Thanathai Lertpetchpun and Tiantian Feng and Dani Byrd and Shrikanth Narayanan}},
  year      = {{2025}},
  booktitle = {{Interspeech 2025}},
  pages     = {{4648--4652}},
  doi       = {{10.21437/Interspeech.2025-1082}},
  issn      = {{2958-1796}},
}

@INPROCEEDINGS{odwyer2017,
    author={O'Dwyer, Jonny and Flynn, Ronan and Murray, Niall},
    booktitle={2017 IEEE International Conference on Bioinformatics and Biomedicine (BIBM)}, 
    title={Continuous affect prediction using eye gaze and speech}, 
    year={2017},
    volume={},
    number={},
    pages={2001-2007},
    doi={10.1109/BIBM.2017.8217968}
}

@INPROCEEDINGS{lanata2011,
    author={Lanatà, Antonio and Armato, Antonino and Valenza, Gaetano and Scilingo, Enzo Pasquale},
    booktitle={2011 5th International Conference on Pervasive Computing Technologies for Healthcare (PervasiveHealth) and Workshops}, 
    title={Eye tracking and pupil size variation as response to affective stimuli: A preliminary study}, 
    year={2011},
    volume={},
    number={},
    pages={78-84},
    doi={10.4108/icst.pervasivehealth.2011.246056}
}

@article{araluce_2021,
author = {Araluce, Javier and Bergasa, Luis adn Ocana, Manuel and Lopez-Guillen, Elena and Revenga, Pedro and Arango, J and Perez, Oscar},
title = {Gaze Focalization System for Driving Applications Using OpenFace 2.0 Toolkit with NARMAX Algorithm in Accidental Scenarios},
journal = {Sensors (Basel)},
volume = {21},
number = {18},
}

@ARTICLE{soleymani2012,
    author={Soleymani, Mohammad and Lichtenauer, Jeroen and Pun, Thierry and Pantic, Maja},
    journal={IEEE Transactions on Affective Computing}, 
    title={A Multimodal Database for Affect Recognition and Implicit Tagging}, 
    year={2012},
    volume={3},
    number={1},
    pages={42-55},
    doi={10.1109/T-AFFC.2011.25}
}

@INPROCEEDINGS{edinger2024,
    author={Edinger, Janick and Heck, Melanie and Becker, Christian},
    booktitle={2024 IEEE International Conference on Pervasive Computing and Communications Workshops and other Affiliated Events (PerCom Workshops)}, 
    title={Emotion Prediction Through Eye Tracking in Affective Computing Systems}, 
    year={2024},
    volume={},
    number={},
    pages={326-332},
    doi={10.1109/PerComWorkshops59983.2024.10502422}
}

@misc{dosovitskiy_2021,
      title={An Image is Worth 16x16 Words: Transformers for Image Recognition at Scale}, 
      author={Alexey Dosovitskiy and Lucas Beyer and Alexander Kolesnikov and Dirk Weissenborn and Xiaohua Zhai and Thomas Unterthiner and Mostafa Dehghani and Matthias Minderer and Georg Heigold and Sylvain Gelly and Jakob Uszkoreit and Neil Houlsby},
      year={2021},
      eprint={2010.11929},
      archivePrefix={arXiv},
      primaryClass={cs.CV},
      url={https://arxiv.org/abs/2010.11929}, 
}

@misc{su_2023,
      title={RoFormer: Enhanced Transformer with Rotary Position Embedding}, 
      author={Jianlin Su and Yu Lu and Shengfeng Pan and Ahmed Murtadha and Bo Wen and Yunfeng Liu},
      year={2023},
      eprint={2104.09864},
      archivePrefix={arXiv},
      primaryClass={cs.CL},
      url={https://arxiv.org/abs/2104.09864}, 
}

@misc{bengio_2015,
      title={Scheduled Sampling for Sequence Prediction with Recurrent Neural Networks}, 
      author={Samy Bengio and Oriol Vinyals and Navdeep Jaitly and Noam Shazeer},
      year={2015},
      eprint={1506.03099},
      archivePrefix={arXiv},
      primaryClass={cs.LG},
      url={https://arxiv.org/abs/1506.03099}, 
}

@article{huber_1964,
  title={Robust Estimation of a Location Parameter},
  author={Huber, Peter J.},
  journal={Annals of Mathematical Statistics},
  volume={35},
  number={1},
  pages={73--101},
  year={1964},
  publisher={Institute of Mathematical Statistics},
  doi={10.1214/aoms/1177703732}
}

@article{mehrabian_1974,
  author    = {Albert Mehrabian and James A. Russell},
  title     = {An Approach to Environmental Psychology},
  journal   = {MIT Press},
  year      = {1974}
}

@ARTICLE{wang_2018,
  title     = "Automatic emotion perception using eye movement information for
               {E-Healthcare} systems",
  author    = "Wang, Yang and Lv, Zhao and Zheng, Yongjun",
  journal   = "Sensors (Basel)",
  publisher = "MDPI AG",
  volume    =  18,
  number    =  9,
  pages     = "2826",
  month     =  aug,
  year      =  2018,
  copyright = "https://creativecommons.org/licenses/by/4.0/",
  language  = "en"
}

@article{wu_2021,
title = {Emotion classification on eye-tracking and electroencephalograph fused signals employing deep gradient neural networks},
journal = {Applied Soft Computing},
volume = {110},
pages = {107752},
year = {2021},
issn = {1568-4946},
doi = {https://doi.org/10.1016/j.asoc.2021.107752},
url = {https://www.sciencedirect.com/science/article/pii/S1568494621006736},
author = {Qun Wu and Nilanjan Dey and Fuqian Shi and Rubén González Crespo and R. Simon Sherratt},
}

@ARTICLE{lim_2020,
  title     = "Emotion recognition using eye-tracking: Taxonomy, review and
               current challenges",
  author    = "Lim, Jia Zheng and Mountstephens, James and Teo, Jason",
  journal   = "Sensors (Basel)",
  publisher = "MDPI AG",
  volume    =  20,
  number    =  8,
  pages     = "2384",
  month     =  apr,
  year      =  2020,
  copyright = "https://creativecommons.org/licenses/by/4.0/",
  language  = "en"
}

@misc{avramidis_2025,
      title={Deep Learning Characterizes Depression and Suicidal Ideation from Eye Movements}, 
      author={Kleanthis Avramidis and Woojae Jeong and Aditya Kommineni and Sudarsana R. Kadiri and Marcus Ma and Colin McDaniel and Myzelle Hughes and Thomas McGee and Elsi Kaiser and Dani Byrd and Assal Habibi and B. Rael Cahn and Idan A. Blank and Kristina Lerman and Takfarinas Medani and Richard M. Leahy and Shrikanth Narayanan},
      year={2025},
      eprint={2504.20944},
      archivePrefix={arXiv},
      primaryClass={cs.LG},
      url={https://arxiv.org/abs/2504.20944}, 
}

@ARTICLE{sarkar2022,
  author={Sarkar, Pritam and Etemad, Ali},
  journal={IEEE Transactions on Affective Computing}, 
  title={Self-Supervised ECG Representation Learning for Emotion Recognition}, 
  year={2022},
  volume={13},
  number={3},
  pages={1541-1554},
  doi={10.1109/TAFFC.2020.3014842}}

@ARTICLE{wu2024,
  author={Wu, Yujin and Daoudi, Mohamed and Amad, Ali},
  journal={IEEE Transactions on Affective Computing}, 
  title={Transformer-Based Self-Supervised Multimodal Representation Learning for Wearable Emotion Recognition}, 
  year={2024},
  volume={15},
  number={1},
  pages={157-172},
  doi={10.1109/TAFFC.2023.3263907}}

@ARTICLE{sun2025,
  author={Sun, Licai and Lian, Zheng and Wang, Kexin and He, Yu and Xu, Mingyu and Sun, Haiyang and Liu, Bin and Tao, Jianhua},
  journal={IEEE Transactions on Affective Computing}, 
  title={SVFAP: Self-Supervised Video Facial Affect Perceiver}, 
  year={2025},
  volume={16},
  number={1},
  pages={405-422},
  doi={10.1109/TAFFC.2024.3436913}}

@misc{nimitsurachat2023,
      title={Self-Supervised Learning for Audio-Based Emotion Recognition}, 
      author={Peranut Nimitsurachat and Peter Washington},
      year={2023},
      eprint={2307.12343},
      archivePrefix={arXiv},
      primaryClass={cs.SD},
      url={https://arxiv.org/abs/2307.12343}, 
}

@inproceedings{loshchilov_2019,
  title     = {Decoupled Weight Decay Regularization},
  author    = {Loshchilov, Ilya and Hutter, Frank},
  booktitle = {Proceedings of the 7th International Conference on Learning Representations (ICLR)},
  year      = {2019},
  url       = {https://arxiv.org/abs/1711.05101}
}

\end{document}